\pdfoutput=1

\documentclass[11pt]{article}
\usepackage{microtype}
\usepackage{acl}
\usepackage{soul}
\usepackage{bm}
\usepackage{amsmath}
\usepackage{amssymb}
\usepackage{times}
\usepackage{latexsym}
\usepackage{graphicx}
\usepackage{subfigure}

\usepackage{caption}
\usepackage{mathrsfs}
\usepackage[T1]{fontenc}

\usepackage[utf8]{inputenc}

\usepackage{microtype}

\usepackage{graphicx}

\usepackage{multirow}
\usepackage{enumitem}

\usepackage[T1]{tipa}
\usepackage{makecell}

\usepackage{url}
\usepackage{tablefootnote}

\title{Smells like Teen Spirit:\\ An Exploration of Sensorial Style in Literary Genres}

\author{Osama Khalid, Padmini Srinivasan \\
  University of Iowa\\
  \texttt{\{osama-khalid, padmini-srinivasan\}@uiowa.edu} \\}

\begin{document}

\maketitle

\begin{abstract}
\vspace*{-0.5em}
It is well recognized that sensory perceptions and language have interconnections through
numerous studies in psychology, neuroscience, and sensorial linguistics. Set in this rich context we ask whether the use of sensorial language in writings is part of linguistic style? This question is important from the view of stylometrics research where a rich set of language features have been explored, but with insufficient attention given to features related to sensorial language. 
Taking this as the goal we explore several angles about sensorial language and style in collections of lyrics, novels, and poetry.  We find, for example, that individual use of sensorial language is not a random phenomenon; choice is likely involved. Also, sensorial style is generally stable over time - the shifts are extremely small.
Moreover, style can be extracted from just a few hundred sentences that have sensorial terms. We also identify representative and distinctive features within each genre.
For example, we observe that 4 of the top 6 representative features in novels collection involved individuals using olfactory language where we expected them to use non-olfactory language.



\end{abstract}

\section{Introduction}

Sensory perceptions shape how we use language and communicate \cite{paradis2003notion}.
When we use sensorial words (i.e. words with meanings connected to our senses) like \textit{fuzzy} or \textit{stinky}, besides communicating sensorial experiences these also stimulate perceptual systems in the recipient's mind \cite{speed2020grounding}.

The space of senses -- sometimes called the ``Aristotelian'' senses \cite{sorabji1971aristotle},  include the five modalities: visual, auditory, haptic, gustatory, and olfactory. 
Relatively recently, linguistics and psychologists have added a sixth sense -- interoception \cite{craig2002you}. This refers to the perception of sensations from inside the body, both physical such as hunger and pain, and emotional, such as joy. 
This sensory space 
has been the basis of much prior research.

\noindent\textbf{Sensorial Linguistics} is about studying how language relates to the senses.
A key focus has been to study how different sensorial experiences and perceptions are packaged into linguistic units \cite{winter2019sensory}.
Researchers have looked at how some senses dominate in language \cite{wintervision}, how sensorial language varies across lexical categories \cite{lievers2018sensory} and how sensory experiences influence sensorial language \cite{croijmans2019measuring,murphy2019olfactory}. However, the domain of sensorial linguistics is still nascent with many unexplored questions.

\noindent\textbf{Stylometrics:}
As individuals grow, besides consciously learning a vocabulary they also develop a linguistic style. Some stylistic elements may be acquired subconsciously and others by choice.
Stylistic choices can reflect the individual's social reality or affective state \cite{savoy2020machine}.
 Several categories of stylistic features have been identified, such as the use of function words and language complexity \cite{holmes1998evolution}. Stylometrics is important for goals such as author attribution and affect classification. 

\noindent\textbf{Style in sensorial language use:}
A key limitation in stylometrics is that linguistic style around sensorial language has not been studied systematically. 
A person who wants to express being depressed has several word choices. She could use ``sad'' or the less frequent ``downcast''. Her propensity to choose one or the other may be considered as part of her linguistic style.

Consider a cloudy scene. A person may use visual language focusing on color, and say ``\textit{the clouds are white}''. Another may use haptic language focusing on texture, ``\textit{the clouds are fluffy}''. 
While sensorial language is clearly important for communication, we do not yet know if there are distinguishable patterns in sensory language use at the level of individuals, texts, etc. This gap in stylometrics motivates us to ask the following about sensorial language style:

\begin{itemize}[noitemsep,topsep=0pt,leftmargin=*]
    \item \textbf{RQ 1: }Is the notion of sensorial style meaningful or is it a product of random chance?
    \item \textbf{RQ 2:} How much data do we need to get a stable representation of sensorial style?
    \item \textcolor{black}{\textbf{RQ 3: }Does sensorial style vary with time?}
    \item \textcolor{black}{\textbf{RQ 4:} Which features are representative and distinctive of the individuals within each genre?} 
\end{itemize}

\section{Representing Sensorial Language Style}\label{methods:alpha}
\vspace{-0.5em}
Sensorial style may be represented at different levels of abstractions. At the lowest level, we can represent the proportion of an individual's language that is sensorial and also examine the frequencies of different sensorial words. At a higher level of abstraction, we can ask how frequent are different sense modalities (visual, auditory, etc.) in an individual's language.  Alternatively, we can represent style by the extent to which a person's use of sensory modalities aligns with general expectations. This is related to synaesthesia, where one sense modality is used when another is expected - a well studied phenomenon in sensorial linguistics \cite{lievers2015synaesthesia,de1945romanticism}. As an example in their work, \citeauthor{lievers-huang-2016-lexicon} (\citeyear{lievers-huang-2016-lexicon}) developed a lexicon of perception that they used to automatically identify perception related synaesthetic metaphors. 
Similarly we also approach the problem of sensorial style through the lens of their synaesthetic usage.   

In the 2010 animated film `Despicable Me', the character of Agnes hugs a unicorn and says ``It is so \textit{fluffy}''. One reason why this quote acquired somewhat of a meme status is because it subverted audience expectations of a more visual word like ``\textit{pretty}'' or ``\textit{white}'' to describe the unicorn.
Instead she opts for the more unexpected haptic word ``\textit{fluffy}''. This substitution of visual language for haptic, a synaesthesia, might indicate that Agnes' perceives the world in a more tactile manner rather than in a visual way. 
In order to assess if this is a stylistic tendency, we can examine all of Agnes' language use and ask the general question: to what extent does she use haptic language in contexts where we generally expect visual language? We can ask similar questions related to each combination of expected versus observed sensory modalities.
Observations for all combinations, including the homogeneous non-synaesthesia ones, are accumulated to form Agnes' (or any other individual's, group's or genre's) sensory style representation.

\noindent
\textbf{Terminology and notation:} More formally, 
 we consider a sentence to be a ``sensorial sentence'' if it has at least one word or phrase that appears in a sensorial lexicon. 
Further, we define a ``sense-focused sentence'' to be a sensorial sentence with a single sensorial term selected as focus term. Thus, if a sensorial sentence has $n$ sensorial terms then we derive from it $n$ sense-focused sentences. 

\vspace{-0.2em}

Assume $S_i=\{S_{i1},S_{i2},...,S_{in}\}$ is the set of $n$ sense-focused sentences identified from the writings of individual $i \in I$.
Let $C=\{H,V,I,O,G,A\}$ represent the modalities: Haptic, Visual, Interoceptive, Olfactory, Gustatory, Auditory respectively. 
Using $\bar{N}$ to  represent concepts that are ``not-sensorial'' we define $\widetilde{C}$ as: 
$\widetilde{C}=C \cup \{\bar{N}\}$. 

We also define two functions.
$F(\hat{s}_{ij})$ is a sensory lexicon lookup function that returns the sensorial category $c \in C$ for the focused sensorial term $\hat{s}_{ij}$ in the sense focused sentence $S_{ij}$. 
E.g., given the sensorial sentence ``\textit{The unicorn is \textbf{white} and \textbf{fluffy}}'', we have two sense-focused sentences $S_{ij}$ and $S_{ik}$ corresponding to the two sensorial terms, $\hat{s}_{ij}$ and $\hat{s}_{ik}$. For $S_{ij}$ with the focus term ``\textit{\textbf{white}}'' $F(\hat{s}_{ij})$ will return $V$. 
For $S_{ik}$ with focus term ``\textit{\textbf{fluffy}}'' it will return $H$.
\vspace{-0.3em}

The second function we define is $M(S_{ij})$ which returns the  ``expected''  modality $c\in\widetilde{C}$ for the same focus term in $S_{ij}$

We describe this function in Section \ref{method:expectation}.

\noindent
\textbf{Calculating observed to expected ratios:}
We represent an individual's sense-focused sentences as a list of length $|S_i|$. Each entry is a pair of expected and observed modalities of the form $[(M(S_{ij}),(F(\hat{s}_{ij})]$.
For observed modality $y \in C$ and expected modality $x \in \widetilde{C}$, the observed to expected ratio $\alpha_{i}^{xy}$ for individual $i$ is:
\setlength{\abovedisplayskip}{3pt}
\setlength{\belowdisplayskip}{3pt}
\begin{equation}\label{eq:alpha}
\alpha_{i}^{xy}=\frac{\left|\{S_{it}:F(\hat{s}_{it}) = y \text{ and } M(S_{it})=x\}\right|}{\left|\{S_{it}: M(S_{it})=x\}\right|}
\end{equation}

Note that these ratios are the informative units of sensorial style. For example, if the ratio is $1$ when $x$ and $y$  are the same modality, a homogeneous combination, then the individual's use of that modality is highly aligned with general expectation.
On the other hand, if it is close to $0$ then she deviates considerably from the expected use of modality $x$, i.e. there is greater synaesthesia.

\noindent
\textbf{Style vectors:}
For each $x \in \widetilde{C}$, we then concatenate its $6$ ratios into a vector of the form:
\setlength{\abovedisplayskip}{3pt}
\setlength{\belowdisplayskip}{3pt}
\begin{equation}\label{eq:expectedvector}
s_{i}^{x}=\underset{y\in C}{\sqcup}\alpha_{i}^{xy}
\end{equation}
It follows that 
\setlength{\abovedisplayskip}{3pt}
\setlength{\belowdisplayskip}{3pt}
\[
    \sum_{y \in C} \alpha_{i}^{xy}= 
\begin{cases}
    1,\quad& \text{if } |\{S_{it}: M(S_{it})=x \}| \geq 1\\
    0,\quad              & \text{otherwise}
\end{cases}
\]

We can now define the sensorial style vector $u_i$ of  $i \in I$ as a concatenation of the seven vectors, one for each expected modality.
The size of $u_i$ is $x \times y=42$.
\begin{equation}\label{eq:vectorRep}
u_{i}=\underset{x\in \widetilde{C}}{\sqcup }s_{i}^{x}
\end{equation}

\vspace{-1em}
\subsection{Implementing function $M(S_{ij})$}\label{method:expectation}

Given a sense-focused sentence, function $M$ returns the expected modality $c\in \widetilde{C}$ of the sentence's focus sensorial term as per general expectation in English. 
We leverage RoBERTa-MLM\footnote{We experimented with BERT as well, however, RoBERTa gave us more accurate results.} as a stand-in for general English language usage. RoBERTa is a transformer based language model pre-trained on 160 GB of data \cite{liu2019roberta}. Prior works like \cite{mosbach2020closer} and \cite{sinha2020unnatural} have shown that language models learn the norms of the language on which they are trained. 

This makes them ideal for our task. 
We mask the focus sensorial words in our sentences and input them to the model. RoBERTa returns the probabilities for all the words in its vocabulary at each masked location. Probabilities represent likelihood of appearance of the words at that location. We use these probabilities to identify the expected sense modality at each masked location as follows.

Let $W = \{(w_1,p_1),(w_2,p_2)\hdots(w_N,p_N)\}$ 
be the ranked set of words returned by RoBERTa for a masked location (location of focus sensorial term) in $S_{ij}$; top ranked has highest probability.

Using $F(w_k)$, we lookup the sense for each word in the top 100. We combine this information to get an aggregate probability score $\Pi(c,S_{ij})$ for each modality $c$ as follows:
\begin{equation}
\Pi(c,S_{ij})=\underset{\substack{k \leq 100\\ F(w_k)=c }}\sum p_k   
\end{equation}
For greater confidence we only include  $S_{ij}$ in our analysis if its majority modality has $\Pi(c,S_{ij})$  $>0.5$.
We then define
\begin{equation}
    M(S_{ij})=\operatorname*{argmax}_{c\in \widetilde{C}} \Pi(c,S_{ij})
\end{equation}
In essence, $M(S_{ij})$ returns the expected modality with the highest aggregate probability for the focused sense word in $S_{ij}$ as determined using RoBERTa.

\subsection{Sensorial Lexicon}

\begin{table}[h]
\scriptsize
\resizebox{0.9\columnwidth}{!}{%
\begin{tabular}{l||l|l||l|l}
&\multicolumn{4}{c}{Lexicon}   \\\hline
 &\multicolumn{2}{c||}{Original} & \multicolumn{2}{c}{Modified}  \\\hline
 Modality & $N$ & $\%$ & $N$ & $\%$ \\\hline
Visual & 29552 & 75.0 & 9419 & 50.2 \\
Interoceptive & 3546 & 9.0 & 3449 & 18.4 \\
Auditory & 4528 & 11.5 & 3803 & 20.3 \\
Haptic & 675 & 1.7 & 972 & 5.2 \\
Gustatory & 890 & 2.3 & 890 & 4.7 \\
Olfactory & 216 & 0.5 & 216 & 1.2 \\\hline
Total & 39407 & 100 & 18749 & 100
\end{tabular}
}
\caption{Distribution of modalities in original \citeauthor{lynott2020lancaster} (\citeyear{lynott2020lancaster}) lexicon and our modified subset lexicon.}

\label{tab:lexicon}
\end{table}

\vspace{-1em}

We use the sensorimotor norms lexicon published recently \cite{lynott2020lancaster} which has 39,954 concepts from the English Language.
\citeauthor{brysbaert2016many} estimates that the average adult lexicon is composed of approximately $42,000$ words.  Therefore this lexicon approximates a significant majority of everyday English. 

Each concept was rated by annotators along a $0$-$5$ scale for the six modalities (Auditory, Gustatory, Haptic, Olfactory, Visual, Interoceptive)
For example ``\textit{fluffy}'' is rated $4.41$ for Haptic, $0.29$ for Gustatory,  $3.77$ for Visual, $0.35$ for Interoceptive and $0$ for Auditory and Olfactory. 

 The dimension with the highest rating is the dominant modality. 

Dominance alone is not enough to ensure that a concept belongs to a particular sensorial modality since almost half of the concepts score less than $2.55$ on any sense modality. 
Therefore,
we filter the lexicon  
by ranking all concepts in a given modality by their rating and selecting only those in the top quartile. This ensures strong alignment to dominant modalities. 
Table \ref{tab:lexicon} describes the lexicons\footnote{There are other sensory lexicons like \cite{lievers2016lexicon} and \cite{lynott2013modality}. Besides being the largest and most recent the \citeauthor{lynott2020lancaster} (\citeyear{lynott2020lancaster}) sensorimotor lexicon is the only one to include ratings for interoception.}

\section{Methods}

\subsection{RQ 1: Is Sensorial Style a Product of Random Chance?}\label{method:randomVec}
 In order to be meaningful our  representation of sensorial style should not be a product of random chance. 

If an individual chooses sensorial modalities randomly and not deliberately then we expect her observed and expected modality distributions to be independent of each other. 

For example in  ``\textit{the clouds are \textbf{white}}'' the expected modality may be visual but the individual randomly chooses from one of the six senses. We use this random model in our analysis.

\textcolor{black}{As a first step, for any $i \in I$ with a set of $n$ sense-focused sentences $S_i$, 
we define, $\Gamma(i)$, the distribution of the sense modalities in $C$ observed in $S_i$. For each sense-focused term $\hat{s_{ij}}$ in $S_i$, we use a function $\bar{F}(\hat{s_{ij}},\Gamma(i))$ that returns a random modality $c \in C$ with distribution $\Gamma(i)$.}

\textcolor{black}{For each $i \in I$, we create $m$ random pseudo-documents $\mathcal{R}_{i}=\{\mathcal{R}_{i}^{1},\mathcal{R}_{i}^{2}\hdots \mathcal{R}_{i}^{m}\}$. Each random pseudo-document has the same set of sense-focused sentences $S_i$. However, instead of using $F(\hat{s}_{ij})$ to look up the modality, we use $\bar{F}(\hat{s}_{ij},\Gamma(i))$ to get a random modality. 
Equation~\ref{eq:alphaN}, a modification of equation \ref{eq:alpha}, gives us $\bar{\alpha}_{ik}^{xy}$ which is used to calculate the style vector $u_i^k$ for random pseudo-document $\mathcal{R}_i^k$. }
\vspace{-0.4em}
\begin{equation}\label{eq:alphaN}
\bar{\alpha}_{ik}^{xy}=\frac{\left|\{S_{ij}:\bar{F}(\hat{s}_{ij},\Gamma(i)) = y \text{ and } M(S_{ij})=x\}\right|}{\left|\{S_{ij}: M(S_{ij})=x\}\right|}
\end{equation}

\textcolor{black}{Thus, for each $i \in I$ with sensorial style vector $u_i$, we have $m$ random style vectors $\{u_i^1,u_i^2 \hdots u_i^m\}$ generated from the random pseudo-documents in $\mathcal{R}_i$.}

Let $U_i=\{u_i\} \cup \{u_i^1,u_i^2\hdots u_i^m\}$. For each vector $v \in U_i$, we calculate its average cosine similarity with all other elements in $U_i$.
Ranking the elements of $U_i$ by decreasing order of average similarity we check whether the style vector $u_i \in U_i$ 
has lower average similarity than at least $95\%$ of the vectors in $U_i$ (i.e. $p$-value $<0.05$ ). 
If so, we infer with $95\%$ confidence that $i$'s style vector, $u_i$, is not random and therefore likely a product of an individual stylistic choice.

\vspace{-0.5em}
\subsection{RQ 2: How much data is needed to describe sensorial style?}\label{method:selfSim}
\vspace{-0.3em}

Given the set of sense focused sentences 
$S_i$, where $|S_{i}|=n$, we randomly sample subsets from $S_i$ of size $k$ and compute the style vector from each sample. We explore how increasing the values for $k$ affect style convergence.

For a given sentence set size $k$, we identify $m$ random samples (with replacement) of $S_i$, each of size $k$. Thus, we create a set of sentence sets,  $T_i^k=\{\hat{T}_{i}^{k1},\hat{T}_{i}^{k2},\hdots,\hat{T}_{ki}^{m}\}$. For each sentence set $\hat{T}_{i}^{kj} \in T_i^k$, we use the method discussed in Section \ref{methods:alpha} to generate the corresponding style vector $\hat{u}^{kj}_i$. This gives us a set of $m$ sensorial style vectors, $\bm{\hat{u}^k_i}=\{\hat{u}^{k1}_i,\hat{u}^{k2}_i,\hdots,\hat{u}^{km}_i\}$. We then use cosine similarity to calculate the average pairwise similarities between all elements in $\bm{\hat{u}^k_i}$. We recompute this average self-similarity $\overline{\text{sim}(\bm{\hat{u}^k_i})}$ for different values of $k$ in increasing steps of $r$.
We say that the style of the individual has converged for a minimum of $k$ sensorial sentences if $ \overline{\text{sim}(\bm{\hat{u}^k_i})} \approx \overline{\text{sim}(\bm{\hat{u}^{k+r}_i})}$, where $k+r$ is the next sentence set size tested.

\vspace{-0.8em}

\subsection{RQ 3: Does sensorial style vary over time?}\label{sec:temporal}

Here we investigate whether style vectors evolve over time spans.
We segment the writings by time and consider how the average similarity in style varies with temporal distance. 
We first identify all pairs of time points $t_a$ and $t_b$ that are $\gamma$ duration apart.
We then build style vectors for each author with text anchored at $t_a$ and for each author with text anchored at $t_b$.
We then compute the average pairwise similarity between $t_a$ and $t_b$ style vectors.
We repeat this for all values of $\gamma$ that are of interest.

We use a notion of windowing around the time points ($t_a$ and $t_b$) to reduce noise. Each window is of size $\delta$ and distributed equally around each time point. For example, for $t_a$ we create an individual's style vector from all the sense-focused sentences that were published in the range  $t_a-\frac{\delta}{2}<\tau_a<t_a+\frac{\delta}{2}$.

\subsection{\textcolor{black}{RQ 4: Which features are representative and distinctive of the individuals within each genre?}}\label{sec:distRep}
\vspace{-0.5em}
\textcolor{black}{A genre can be represented by the set of sensorial style vectors of its members. Each style vector is composed of 42 features that explore synaesthesia. We are interested in exploring which features are representative of the members of a genre, and also features that make the members distinct.}

\textcolor{black}{We consider a sensorial style feature to be representative if the variation in its usage is low. 
This would indicate that the members use the feature in a consistent manner. 
Formally, a stylistic feature $\alpha^{p}$ is more representative for the members of a genre than another feature $\alpha^{q}$ if its standard deviation $\sigma(\alpha^{p})$, across all the members is lower than the standard deviation $\sigma(\alpha^{q})$.}
At the other end, a high variation would indicate that the feature is distinctive amongst the members.

\vspace{-1em}
\section{Datasets}

\begin{table}[]
\centering
\resizebox{\columnwidth}{!}{%

\begin{tabular}{l||l|l|l|l||l}
\textbf{Genre}&\textbf{\# Authors}&\textbf{\# Works}&\textbf{\#Sentences}&\makecell{\textbf{\# Sensorial}\\ \textbf{Sentences}}&\makecell{\textbf{\# Sensorial}\\ \textbf{Expressions}}\\\hline
Novels & 130 & 317 &1,525,894& 156,570 (10\%)& 474,299 \\
Lyrics & 5,321 & 20,785&1,007,090 & 754,572 (75\%)& 1,501,501\\
Poetry & 1,246 & 3,315&85,236 & 4,979(6\%)& 8,209\\
\end{tabular}
}
\caption{Dataset details for each genre. The percentage of total sentences that are sensorial is in parentheses.}
\label{tab:summaryData}
\end{table}

We analyze 3 literary genres --- novels, poems, music lyrics. 
Compared to poems and lyrics, novels can span tens of thousands of sentences. 
Additionally, novels and poetry are generally associated with a single author. 
Lyrics are sometimes collaborations, however, we assume an artist would not perform a song that is in a style they do not like. Thus, we assume music lyrics to be a reflection of the artist's style. 

\noindent\textbf{Novels: }
We collected English language novels from the Domestic fiction genre of Project Gutenberg\footnote{https://www.gutenberg.org/}. There were 317 works written by 130 authors, with the earliest by Henry Fielding from the early 18$^{\text{th}}$ century and the latest by Rebecca West from the mid-20$^{\text{th}}$ century.

\noindent\textbf{Lyrics: }
We collected songs that were listed on the Billboard Hot 100 charts, 1963 to 2021 (inclusive). This weekly chart ranking  of song popularity is considered the industry standard \cite{whitburn2010billboard}. We assume the first time a song is listed to be its year of production.
We obtained song lyrics using the Genius API\footnote{Some lyrics were not available on http://www.genius.com, because they were instrumentals like the ``\textit{Star Wars Theme}'' which hit No. 1 in 1977, or were not in the Genius database.}. There are $20,785$ song lyrics.

\noindent\textbf{Poetry:} 
Following works like \cite{lou2015multilabel}, we used the corpus of poems available on the Poetry Foundation's\footnote{The Poetry Foundation was established in 2003, and one of its goals is to make ``the best poetry'' accessible (https://www.poetryfoundation.org/foundation/about).} website. To make this dataset more comparable to the lyrics dataset, we only included works published after 1963.

\vspace{-0.8em}
\section{Results}
\begin{table}[!h]
\footnotesize
\centering
\begin{tabular}{l|ll}
 Genre& $>95^{th}$ &\textit{N}\\\hline
Novels & 112 &123 \\
Lyrics & 701 &735 \\
Poetry & 20 &85\\
\end{tabular}

\caption{Number of individuals with lower average similarity than $95\%$ of random vectors. }
\label{tab:randomSimilarity}
\end{table}
\vspace{-1em}
We present our results in two parts. First, we make our general observations. Second, we present results related to our specific research questions.

\vspace{-0.8em}
\subsection{General Observations}

\begin{figure}[h]
\hspace{-0.75cm}
\centering
\includegraphics[width=0.9\columnwidth]{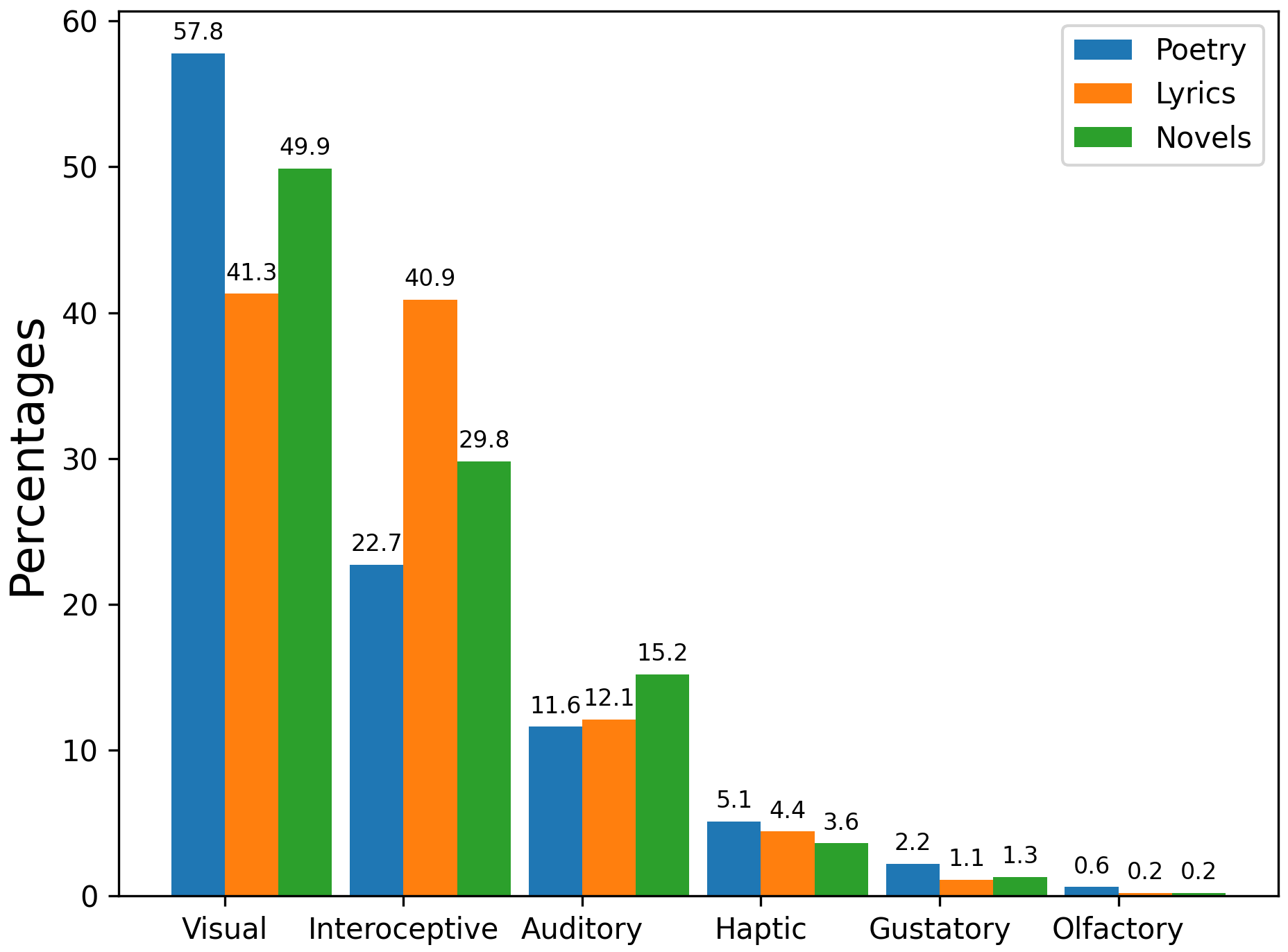}
\caption{Distribution of observed modalities.}
\label{fig:distribution}
\end{figure}

\begin{table*}[!htb]
    
    \begin{minipage}{1.05\columnwidth}
      \centering
      \scriptsize
        \begin{tabular}{l|l|l|l}
            &Novels  & Lyrics  & Poetry\\\hline
            &$I-O$  &    $I-O$  &$A-G$\\
            &$\bar{N}-O$ & $\bar{N}-O$ &   $A-O$ \\
            &$A-O$ &$V-O$ &   $G-I$ \\
            &$I-G$&$A-O$  & $H-O$\\
            &$V-O$ & $I-G$ &$O-A$\\
            &$A-G$  & $V-G$ & $O-H$ \\\hline
            \textbf{Range}&{\tiny(0.00,0.01)}&{\tiny(0.00,0.02)}&{\tiny(0.00,0.00)}\\
            
            \end{tabular}
            \label{tab:rep}
            \caption{The top representative features for each genre.}

    \end{minipage} 
    \begin{minipage}{1.05\columnwidth}
      
      \centering
      \scriptsize
            \begin{tabular}{l|l|l|l}
            &Novels  & Lyrics  & Poetry \\\hline
            &$G-G$ & $H-H$ & $A-A$\\
            &$O-O$ &  $G-G$ & $I-I$ \\
            &$H-H$  &  $A-A$ & $H-H$ \\
            &$G-V$   & $H-V$   & $\bar{N}-V$  \\
            &$H-V$  & $G-V$ & $G-G$ \\
            &$A-A$ & $A-I$ &  $\bar{N}-I$ \\\hline
            \textbf{Range}&{\tiny(0.43,0.15)}&{\tiny(0.44,0.19)}&{\tiny(0.47,0.27)}\\
            \end{tabular}
            \label{tab:dist}
            \caption{The top distinctive features for each genre.}
    \end{minipage}%

    \caption*{Each feature is an (expected,observed) pair e.g., $I-O$ in means that we observe Olfactory language when we expected Interoceptive language. This is the most representative feature for Novels and Lyrics as it has the lowest standard deviation. We include the range of variances of the top most distinctive and representative features.  
    }
\end{table*}
\vspace{-0.5em}
\noindent\textbf{Domination over lower senses: }
The five Aristotelian sensorial modalities have classically been thought of as part of a hierarchy with vision and audition dominating over the three so-called ``lower senses''---Touch, Taste, Smell \cite{howes2010sensual}\footnote{Note interoception is generally not considered in discussions of this hierarchy, possibly because of its relatively recent inclusion \cite{connell2018interoception}.}. This hierarchy manifests in the frequency of language use with the visual and auditory modalities being used more often than the lower senses \cite{majid2018differential}. 
We have consistent results. 
Figure \ref{fig:distribution} shows that for all three genres visual and auditory dominate over the ``lower senses''. Concepts associated with haptic, gustatory and olfactory modalities --- combined, form less than $10\%$ of the total sensorial language. 

While auditory dominates the ``lower senses'' in all three genres, it occurs less than half as often as visual. Going beyond the classical five senses, in all cases interoception dominates the three ``lower senses'' surpassing audition in this regard. Additionally, in lyrics, interoception is as common as visual. 
Clearly interoception with its emphasis on sensations within the body, both physical and emotional, is important in language.



\begin{figure}[htp]
\hspace*{-2.1cm}

    \includegraphics[width=\columnwidth]{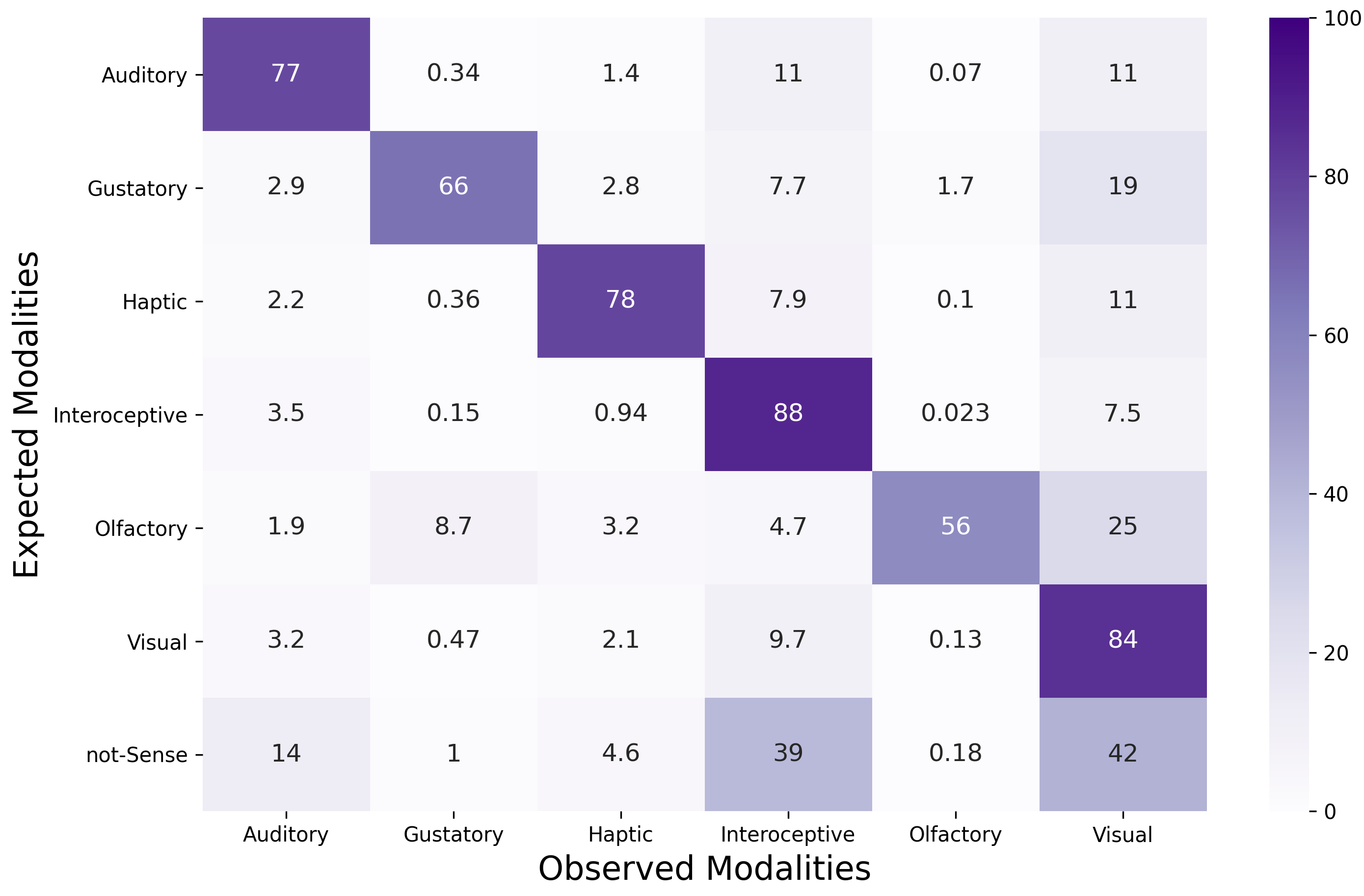}
    \caption{Distribution of expected-observed modalities in Lyrics. Note that we calculate proportions using equation \ref{eq:alpha}, however, for illustrative purposes we show sensorial distribution as percentages.}
    \label{fig:confusion}
\end{figure}

\noindent\textbf{Sensorial style across genres:} 
We investigate how sensorial style varies across genres.  
Using the method described in Section \ref{methods:alpha} we calculate genre-level sensorial style vectors by combining sentence sets at the genre level. We show the distribution over 42 sensorial combinations in Figure \ref{fig:confusion} for just Lyrics. 
Each cell represents an expected-observed modality combination.

\vspace{-1em}

\noindent\textbf{Observed modalities are largely as expected  with some exceptions: }
The diagonal values which are in the range 56 to 88\% indicate that the observed modalities are generally consistent with expected modalities. 
That is, the individuals in our datasets select from the 6 sensorial modalities in a manner that is consistent with the general norms of language use.
The highest consistency is for interoceptive and the lowest is for olfactory. We observe this trend across all genres (see appendix).

Looking at off diagonal values, we observe visual language used in $25\%$ of the cases where we expected olfactory language, $19\%$ and $11\%$ of the cases where we expected gustatory and haptic language, respectively. This usage of visual language as a replacement for lower senses was observed across all 3 genres. This replacement or \emph{cross-modal compensation} might be because the lower senses do not have a strong relation with the perceptual system and consequently individuals might be relying on visual language as a semantic scaffold to compensate for the weaker perceptual system of the lower senses \cite{speed2020grounding}.  

\vspace{-0.2em}
We also observe (in all three genres) 
that in more than $90\%$ of instances where we expected to see non-sensorial language we instead observed interoception. This might also be because interoception dominates in our data and is consistent with observations about higher senses in the literature \cite{majid2018differential}.

\vspace{-1em}
\subsection{Results for research questions}

\textbf{RQ1: Is sensorial style a product of random chance?}
We investigate whether sensorial style is motivated by the individual choices or whether it is a product of randomness. 
 
Table \ref{tab:randomSimilarity} provides the results.
If sensorial style is a non random phenomenon and a product of individual choice and intent, we expect the sensorial style vectors to be distinct from random vectors.
That is, we expect them to have a lower average similarity as compared to random vectors (generated by our random model).  

Methodological details are in Section  \ref{method:randomVec}).

Considering all individuals with more than 10 sense-focused sentences\footnote{Because of the volume of lyrics, we limit the analysis to individuals with $> 500$ sense-focused sentences.}, more than $90\%$ of the individual sensorial style vectors in the novels and lyrics datasets are non-random. 
However, in the poetry dataset only around $23\%$ of the individual vectors in the poetry dataset were non-random. 

A possible reason for the difference in poetry is likely data sparsity --- fewer sensorial expressions/ author (see Table \ref{tab:summaryData}). Exploring this intuition further we find that the non-random vectors in poetry have on average 159 sense-focused sentences while the remaining vectors that looked random had on average 24 sense-focused sentences. Similarly, 
8 out of the 10 most prolific individuals had non-random vectors. However, none of the 10 least prolific individuals had non-random vectors.  These support our intuition regarding data sparsity being the cause of the difference in poetry.

\begin{figure}[!h]
\centering
\includegraphics[width=0.8\columnwidth]{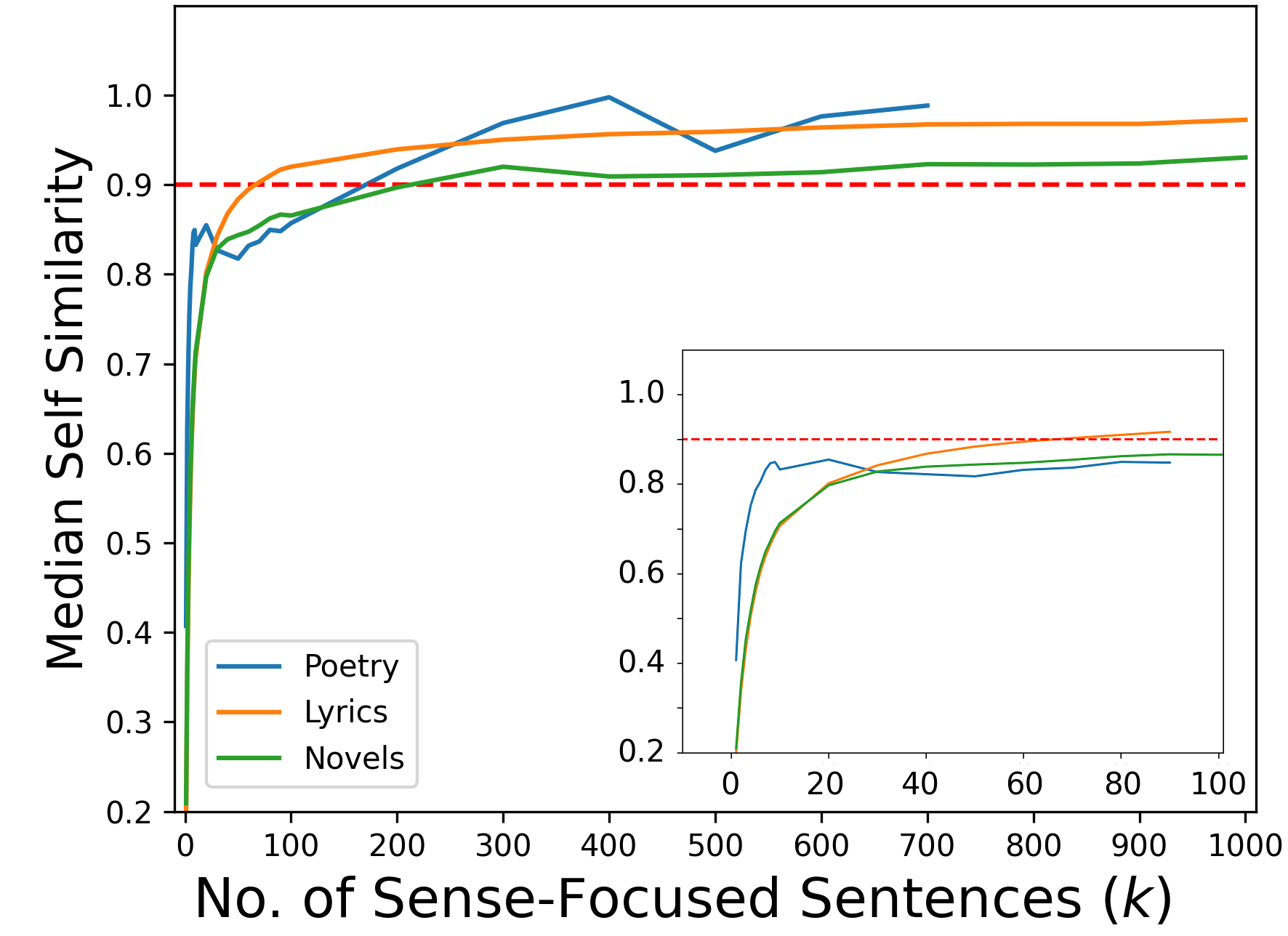}
\caption{Convergence of style vectors as a function of $k$, the number of sense-focused sentences sampled. }
\label{fig:medians}
\end{figure}

\vspace{-0.2em}
\noindent\textbf{RQ2: How much data do we need to get a stable representation of sensorial style?}
We evaluate the average similarity for each individual with progressively larger samples sizes, $k$, of their sense-focused sentences.  We chose a range of values of $k$  from $k=1$ to $k=10$ with a granularity of $1$, from $k=10$ to $k=100$, and $k=100$ to $k=1,000$ with granularities of $10$ and $100$. 
In Figure \ref{fig:medians} we summarize these results with the median of average similarity across all individuals in each genre.
We say the sensorial style vector has converged at a $k$ value when the graph becomes more or less horizontal from that point onward.
From the figure we see that as $k$ increases similarity increases (within the $m$ samples of size $k$). We note that lyrics reaches a median average similarity of $0.90$ with a sample of less than $100$ sense focused sentences. In contrast, we need between $200$ and $300$ sentences to get the same 0.9 median average similarity for novels. Compared to novels and lyrics the plot for poetry has some fluctuations at $k \geq 400$, perhaps because there are only $7$ poets with more than $100$ sentences.

\noindent\textbf{RQ 3: Does sensorial style vary over time?}

\begin{figure}[htp]
\hspace*{-1.1cm}
    \centering
    \includegraphics[width=0.85\columnwidth]{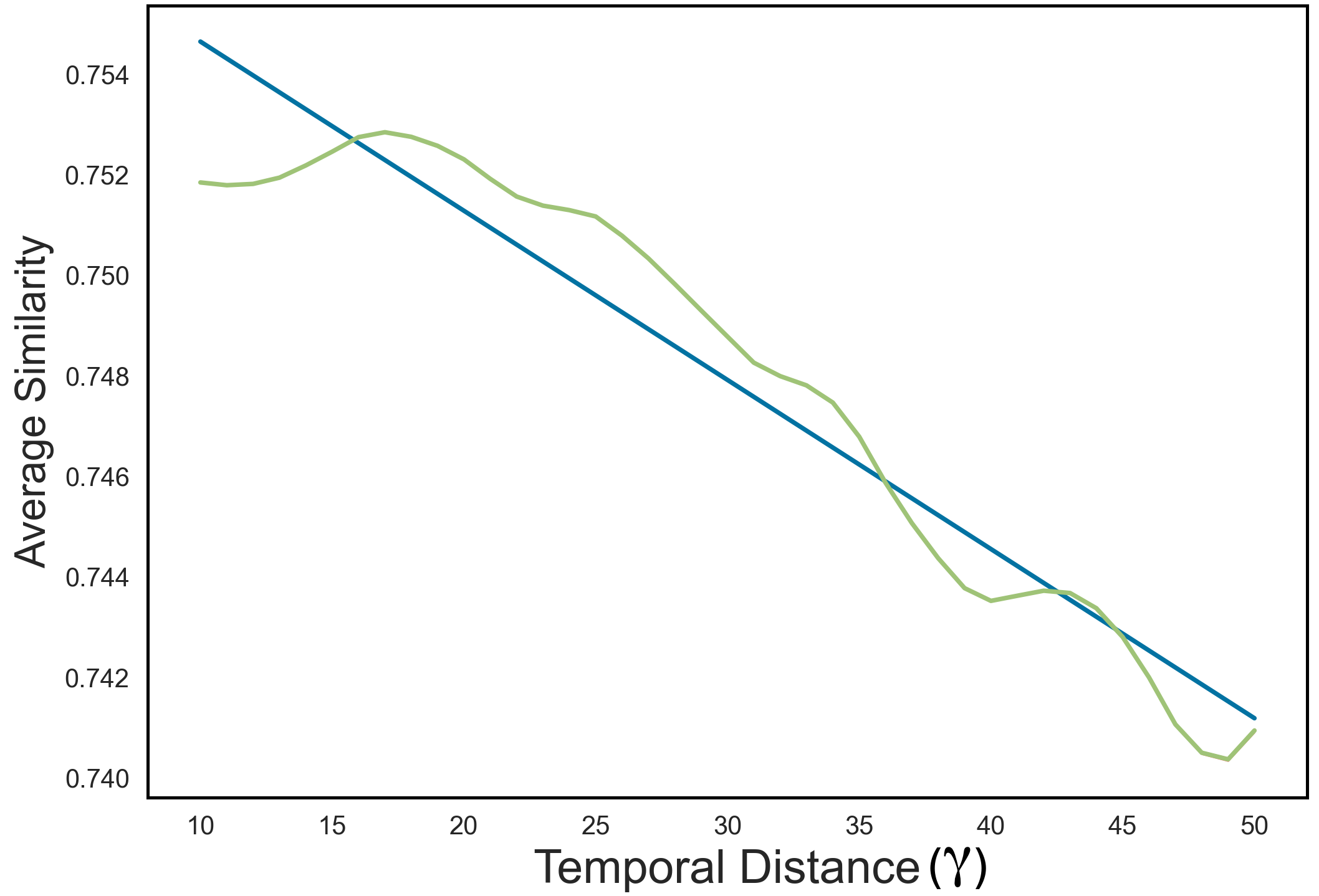}
    \caption{Average similarity between music lyrics as a function of their temporal distance. The blue line represents the linear approximation of the relationship. As temporal distance increases similarity decreases slightly.}
    \label{fig:temporal}
\end{figure}
\vspace{-1em}
We now explore whether sensorial style has changed over time in lyrics.
As observed earlier, the poetry dataset is sparse so we do not include it.
And with novels, it is sometimes challenging to pinpoint a single publication year, since some texts are written over multiple years\footnote{As an example, Louisa May Alcott's \textit{Little Women} was published in two volumes spanning two years, $1868$ and $1869$.}.

Using the method described in Section \ref{sec:temporal}, we measure average similarity in style of lyrics between pairs of individual sensorial style vectors that are a temporal distance $\gamma$ away from each other. We use window length $\delta=1.5$ years.
Figure \ref{fig:temporal} illustrates our findings. 
 
We observe that as temporal distance between works increases the similarity of sensorial style decreases.
The augmented Dickey-Fuller test had a $p-$value $=0.96$ meaning we cannot reject $H_{0}$, that the distribution is not-stationary \cite{cheung1995lag}.
However, the decrease is very slight; the 50 year drop in similarity is only $0.01$.

Approximating the relationship between the average similarity and temporal distance as a linear process, we note that the average similarity decreases very slowly at a rate  of $3.37\times10^{-4}$ per year.

\vspace{-0.1em}
\noindent\textbf{RQ 4: Which features are representative and distinctive of the individuals within each genre?}
\textcolor{black}{Table 4, shows the top-6 features that are representative of the members of each genre. We observe that these are all synaesthetic. Additionally the use of olfactory language in non-olfactory contexts is a representative feature in a majority of table cells (10/18). }
\textcolor{black}{The standard deviations of top representative features is relatively low (between 0.00 to 0.02). This would indicate that the level of consistency for these top features is generally consistent for the three genres.}
One takeaway from these observations can be that in synaesthetic contexts, individuals are more prone to using lower senses (like olfaction) in a more consistent manner.

\textcolor{black}{Table 5, shows the top-6 features that had the highest standard deviation between the members of each genre and were the most distinctive. The distinctive features were predominantly non-syneasthetic. In the cases where the distinctive feature was synaesthetic, the observed modality was either interoceptive or visual. }
This would indicate that there is greater diversity in expressions that rely on higher senses as a semantic scaffold.

For each genre, we rank all the sensorial style features by the standard deviation and compare them using Pearson’s correlation. We observe that, the features are highly correlated. Lyrics had a high correlation with both novels (0.75) and poetry (0.81), while poetry and novels had a slightly lower correlation of 0.48.

\vspace{-0.2em}
\noindent\textbf{Can sensorial style be used for prediction tasks?}

We investigate whether sensorial style features can be used to identify genre. We compare against other standard style representations: LIWC \cite{tausczik2010psychological} and content-free words (CFW) vectors \cite{hughes2012quantitative}.
We use standard 5 fold cross validation for each experiment to train and test a random forest classifier. 
We consider the most prolific $50$ authors/genre. 

\begin{table}[!h]
\footnotesize
\resizebox{\columnwidth}{!}{%
\begin{tabular}{l||l|l|l|l}
Method&Baseline&Sensorial Style&LIWC&CFW\\\hline
Features& --- &42 & 73 & 307\\
Accuracy&0.33&0.91&0.99&0.99\\
\end{tabular}}
\caption{Prediction accuracy of the different features.}
\label{tab:prediction}
\end{table}

Table \ref{tab:prediction} shows the results. We observe that sensorial style predicts genre with a high level of accuracy ($>90\%$).

While the other representations achieve close to perfect accuracies, key to note is that our goal is less about beating baselines and more about understanding the kinds of signals conveyed by sensorial style.

\vspace{-0.5em}
\section{Case Study: Sensorial style in Lyrics}
\vspace{-0.5em}

As a small illustration, we explore how sensorial style varies across  different songs composed by the same artist. 
We consider all $962$ artists who had at least $5$ songs in the Hot 100 and extract a sensorial vector for each song. We then measure the average pair-wise cosine similarity (self similarity) amongst the songs of each artist. 
Almost $80\%$ of the artists had an average self similarity $\geq 0.70$. Only two artists had a self similarity $< 0.50$. 

The rapper \textit{NF} was the most consistent artist with an average self similarity of $0.93$. Conversely, the least consistent artist was the rock musician \textit{Tommy James} with an average similarity of $0.42$. For example, in the song ``When I Grow Up'', \textit{NF} used auditory language non-synaesthetically in $85.7\%$ of the cases and for the song ``NO NAME'' this happened in $76.9\%$ of the cases. The similarity between these two songs was over $95\%$. 

In contrast, ``Nothing to Hide'' and ``Ball and Chain'' by \textit{Tommy James'} had a similarity of $0.32$. In ``Ball and Chain'', the artist uses visual language in all the instances where it was expected. However in ``Nothing to Hide'', he uses interoceptive language synaesthetically instead of visual in about $57\%$ of the cases.
This case study demonstrates a method for exploring sensorial style and their variations across writings at the individual level.

\vspace{-0.7em}
\section{Related Work}
\vspace{-0.7em}
There are no directly comparable studies examining sensorial style for large numbers of individuals that consider interoception. Instead we briefly review closely allied topics.

\noindent
\textbf{Sensorial language:}
Sensorial language is not uniformly distributed across the six sensory modalities as reflected in sensorial lexicons, such as the one we use \cite{lynott2020lancaster}.
This is also observed in large text collections.
In their analysis of 8 million words from around 7,000 English texts \cite{koblet2020online} found over 28,000 visual descriptions and only 78 referring to the olfactory modality.
Similar findings for 
multiple corpora are observed in \cite{winter2018vision}. Our results are consistent with these prior works.

\noindent
\textbf{Sensorial language, the brain \& emotion:}
The salience of sensorial words is known to be highly correlated with the volumes of cortical activation in the brain \cite{reilly2020english}. 
\citeauthor{lievers2015synaesthesia} (\citeyear{lievers2015synaesthesia}) show that there is directionality to how senses are substituted for each other. 

\citeauthor{winter2016taste} (\citeyear{winter2016taste}) found that gustatory and olfactory words (e.g., `stinky', `delicious') are on average more emotionally valenced than visual and auditory words and these also appear in more emotionally valenced sentences. 

\citeauthor{bubl2010seeing} (\citeyear{bubl2010seeing}), show that alterations in mental states have a direct effect on perception; specifically, that depression directly impacts how the color blue was perceived.
\citeauthor{kernot2016impact}
(\citeyear{kernot2016impact}),
found a decrease in the novelist Iris Murdoch's use of olfactory language following her diagnosis of depression and Alzheimer's. We credit this study for providing us with the hint that sensorial language may lead to a sensorial style. 
We take their sensorial style analysis forward with larger collections of authors, several genres and a more informative representation of sensorial style.

\vspace{-0.8em}
\section{Limitations and Conclusions}

\vspace{-0.6em}
We have shown that individuals have sensorial language style and that this sensorial style is a non random phenomenon for novelists and musicians and therefore is likely developed intentionally. 
Interestingly, we also found that it takes just a few hundred sentences to extract stable sensorial style representations.
We also show that sensorial style in lyrics largely stable over time; the average similarity decreases at a rate of $3.37\times 10^{-4}$ per year.

Additionally, we show that sensorial style vectors seem to perform well at genre identification. The performance was high ($>0.90$), however, it was not close to perfect as with other style representations. The question about how sensorial style representations can be improved to increase performance requires further investigation.

Our study is limited in that our method relies heavily on the underlying \citeauthor{lynott2020lancaster} (\citeyear{lynott2020lancaster}) lexicon, and as with similar studies, is only as good as the lexicon.
Additionally, we assume that each term is associated with a single sensorial modality. However, as research in psychology and neurology has shown, sensorial language is cross-modal. We leave this analysis to future work.
In summary, we take a  first step towards showing that sensorial style has a legitimate role in stylometrics research.

\bibliography{reference}
\bibliographystyle{acl_natbib}
\newpage
\appendix
\appendix
\section{Distribution of Modalities}
\begin{figure}[!h]
  \centering
  \begin{minipage}[b]{0.45\textwidth}
    \includegraphics[width=\columnwidth]{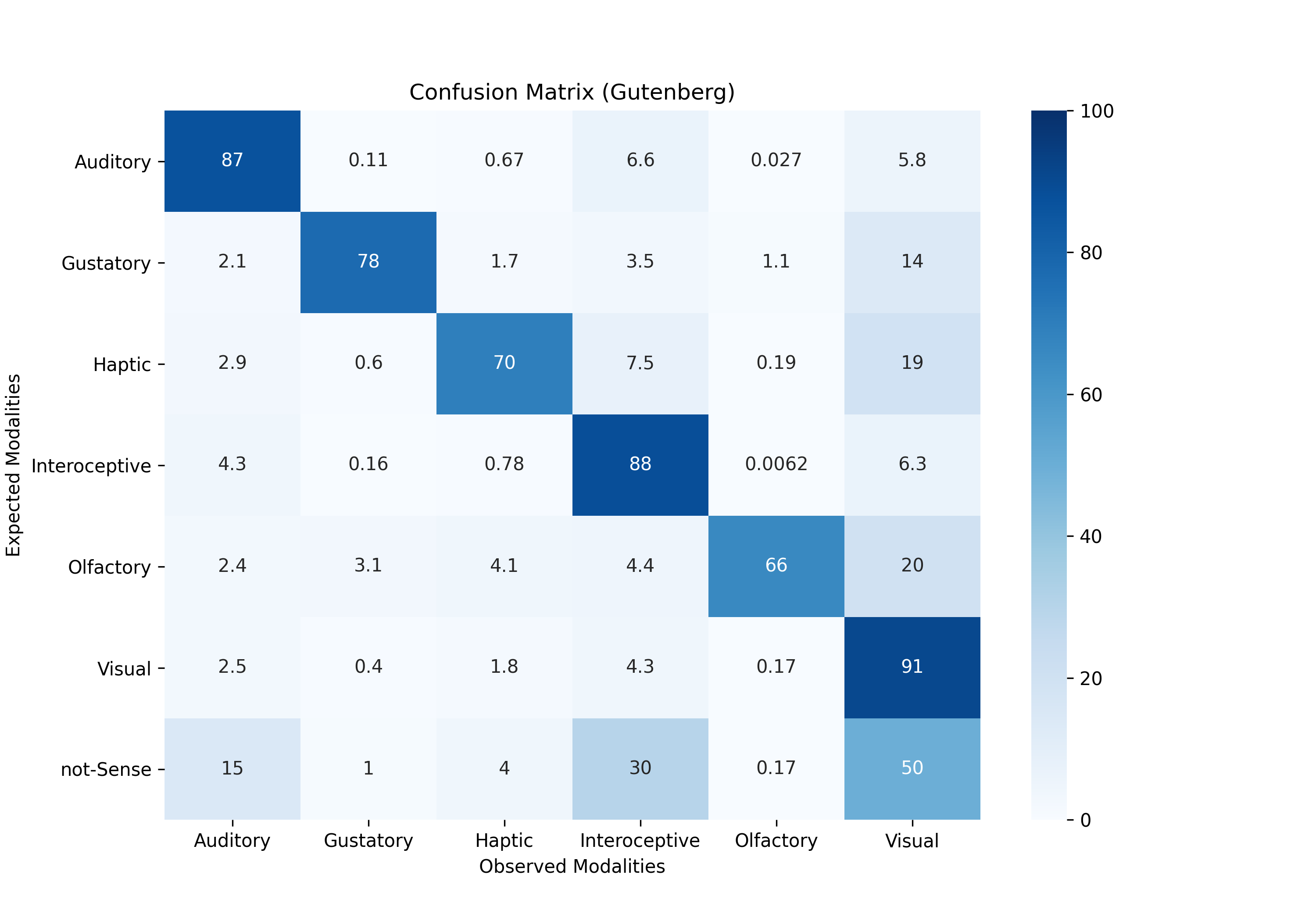}
    \caption{Distribution of expected-observed modalities in the Novels Dataset.}
    \label{fig:confusion_append_guten}
  \end{minipage}
  \hfill
  \begin{minipage}[b]{0.45\textwidth}
    \includegraphics[width=\columnwidth]{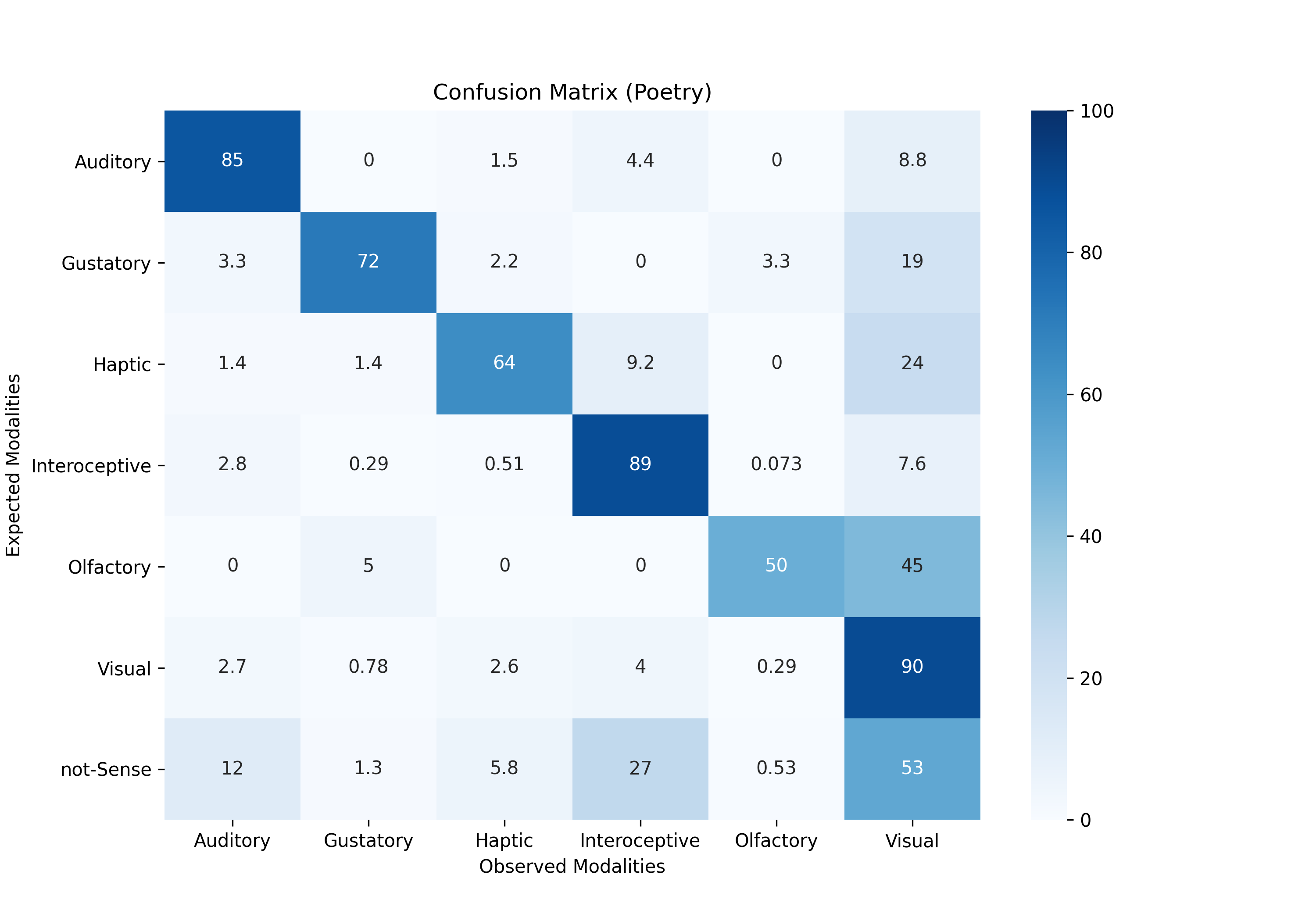}
    \caption{Distribution of expected-observed modalities in the Poetry Dataset.}
    \label{fig:confusion_append_poet}
  \end{minipage}
\end{figure}

\end{document}